\documentclass[lettersize,journal]{IEEEtran}
\usepackage{amsmath,amsfonts}
\usepackage{algorithmic}
\usepackage{array}
\usepackage[caption=false,font=normalsize,labelfont=sf,textfont=sf]{subfig}
\usepackage{textcomp}
\usepackage{stfloats}
\usepackage{url}
\usepackage{verbatim}
\usepackage{graphicx}
\def\BibTeX{{\rm B\kern-.05em{\sc i\kern-.025em b}\kern-.08em
    T\kern-.1667em\lower.7ex\hbox{E}\kern-.125emX}}
\usepackage{balance}

\usepackage{adjustbox}
\usepackage{booktabs}       
\usepackage{amsmath}
\usepackage{multirow}
\usepackage[table]{xcolor}
\usepackage[noadjust]{cite} 
\usepackage{tikz}
\usepackage{collcell}
\usepackage{etoolbox}
\newcommand*{\minvalABXbox}{0.}%
\newcommand*{\midvalABXbox}{6.0}%
\newcommand*{\maxvalABXbox}{50.0}%
\newcommand{\ABXbox}[1]{
    \ifdim #1 pt > \midvalABXbox pt
        \pgfmathparse{max(min(
        100.0*(#1 - \midvalABXbox)/(\maxvalABXbox-\midvalABXbox)
        ,100.0),0.00)} %
        \hspace{-0.33em}
        \xdef\tempa{\pgfmathresult}
        \cellcolor{red!\tempa!white}{#1}
    \else
        \pgfmathparse{max(min(
        100.0*(\midvalABXbox - #1)/(\midvalABXbox-\minvalABXbox)
        ,100.0),0.00)} %
        \hspace{-0.33em}
        \xdef\tempa{\pgfmathresult}
        \cellcolor{black!35!green!\tempa!white!}{#1}
    \fi
 }

\begin{document}
\title{Are discrete units necessary \\ for Spoken Language Modeling?}
\author{Tu Anh Nguyen, Benoit Sagot, Emmanuel Dupoux
\thanks{Tu Anh Nguyen is with Meta and Inria, France, e-mail: nguyentuanh208@gmail.com

Benoit Sagot is with Inria, France, e-mail: benoit.sagot@inria.fr

Emmanuel Dupoux is with Meta and EHESS, ENS-PSL, CNRS, Inria, France, e-mail: emmanuel.dupoux@gmail.com}}


\maketitle

\begin{abstract}
Recent work in spoken language modeling shows the possibility of learning a language unsupervisedly from raw audio without any text labels. The approach relies first on transforming the audio into a sequence of discrete units (or pseudo-text) and then training a language model directly on such pseudo-text. Is such a discrete bottleneck necessary, potentially introducing irreversible errors in the encoding of the speech signal, or could we learn a language model without discrete units at all?
In this work, we study the role of discrete versus continuous representations in spoken language modeling. We show that discretization is indeed essential for good results in spoken language modeling.
We show that discretization removes linguistically irrelevant information from the continuous features, helping to improve language modeling performances. On the basis of this study, we train a language model on the discrete units of the HuBERT features, reaching new state-of-the-art results in the lexical, syntactic and semantic metrics of the Zero Resource Speech Challenge 2021 (Track 1 - Speech Only).
\end{abstract}


\section{Introduction}

Pre-training language models on large-scale text data have achieved tremendous success in 
natural language understanding and
have become a standard in Natural Language Processing (NLP) \cite{radford2018gpt, radford2019gpt2, brown2020gpt3, devlin2018bert, liu2019roberta}.
Recently, \cite{brown2020gpt3} showed that very large language models are actually few-shot learners, and manage to perform well even in zero-shot settings.

Large-scale self-supervised pre-training for speech data has also become more and more popular as a method to boost the performance of Automatic Speech Recognition (ASR) \cite{baevski2020w2v2, weining2021hubert, chung2021w2vbert}.
However, these models mostly rely on fine-tuning, which requires more training and text labels, to either improve the model or evaluate the learned representations of the speech. 
Lately, \cite{nguyen2020zero} introduces a new unsupervised task: Spoken language modeling, the learning of a language unsupervisedly from raw audio without any text labels, along with a suite of 4 zero-shot metrics probing for the quality of the learned models at different linguistic levels: phonetic, lexical, syntactic, semantic.
The metrics are evaluated using the representations extracted from the model (phonetic, semantic) or pseudo-probability scores given by the model (lexical, syntactic). 
Their proposed baseline approach relies on transforming the audio into a sequence of frame-by-frame discrete units (or pseudo-text) and training a language model on the pseudo-text. The trained models displayed better-than-chance performances on nearly all the evaluation metrics of the challenge \cite{nguyen2020zero,dunbar2021zs21}.
However, this paradigm creates a discrete bottleneck between a speech encoder and a language model which could be a potential source of error, and in addition requires multiple training phases (learning an acoustic representation, clustering it, and learning a language model). Is such a discrete bottleneck necessary?

One way in which discrete units could help language modeling stems from the fact that in contrast to text, audio data contains a lot more details, some of which are linguistically relevant (intonation, rhythm, non verbal vocalization), others not so (background noise, reverberation, speaker identity, etc). To the extent that discretization effectively removes linguistically irrelevant information from the continuous features \cite{vanNiekerk2021speakerinfoCPC}, it could indeed help language modeling. Of course, this potential gain could be counterbalanced by the fact that discretization could also make errors and remove useful information.

\begin{figure}[t]
     \centering
     \includegraphics[width=0.5\textwidth]{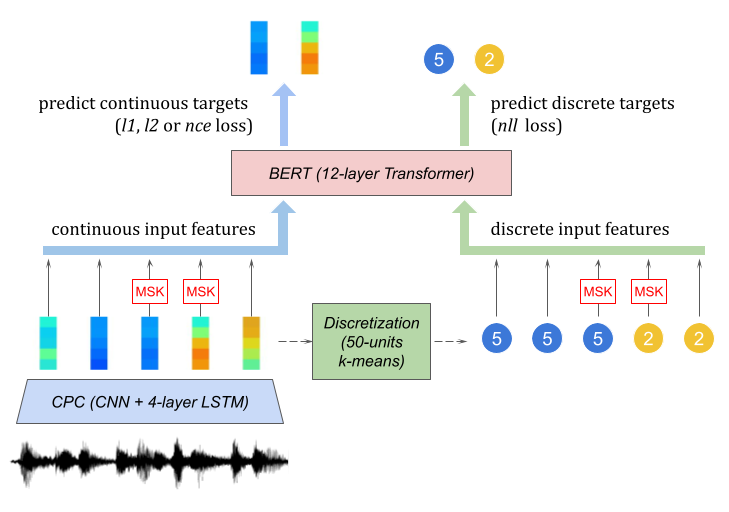}
     \caption{Overview of the trained BERT models. The BERT model takes as input either the continuous features extracted from CPC or the sequences of frame-by-frame discretized units obtained from k-means, and tries to predict either continuous target features (with L1, L2 or NCE loss) or discrete target units (with NLL loss).}
     \label{fig:model}
\end{figure}

In this work,
we analyse the importance of discretization in spoken language modeling.
We employ a pre-trained acoustic model to obtain either continuous or discretized features from audio data. 
We then train BERT language models with a Masked Language Modeling (MLM) objective on both discrete and continuous features used either as inputs or as targets
and evaluate the resulting systems on zero-shot spoken language modeling metrics.
We also evaluate HuBERT \cite{weining2021hubert},
a single model
trained from raw waveform
with discrete targets,
on these metrics and compare the results with our best models.

Our contributions can be listed as follows: 
\begin{itemize}
    \item We show experimentally that discretization is beneficial for spoken language modeling, but we can get rid of discrete bottlenecks by using low-level continuous inputs so long as we still use discrete targets.
    \item We show that discretization disentangles linguistic information from non-linguistic signals, forcing the transformer to focus on linguistic ones.
    \item We show that a self-supervised model trained with a MLM objective on discrete targets like HuBERT achieves very good results on spoken language modeling metrics, showing that it can learn not only acoustic but also high-level linguistic information.
\end{itemize}


\section{Related Work}

\paragraph{Discretization in Self-Supervised Approaches}
Self-supervised models for learning speech representation have become more and more popular as an effective pre-training method for downstream Automatic Speech Recognition (ASR) task, notably wav2vec2.0 \cite{baevski2020w2v2} 
and HuBERT \cite{weining2021hubert}.
Both models comprise a feature extractor (CNN Encoder) followed by a feature encoder (Transformer Encoder), and are trained with a MLM objective like BERT. 
However, wav2vec2.0 discretizes the latent features obtained by the CNN Encoder and uses them as the target for the Transformer Encoder 
using a contrastive loss against negative samples in the sentence. 
On the other hand, HuBERT discretizes fixed features obtained from a teacher model and uses these fixed discrete units as the target for the Transformer Encoder
using a cross-entropy loss.
Finally, our work is mostly similar to \cite{baevski2020effectiveness}, where they compare BERT models training on discrete units obtained from vq-wav2vec \cite{baevski2020vqw2v} and continuous features obtained from wav2vec \cite{scheider2019w2v} on the ASR task. They found that training BERT model on discrete vq-wav2vec units is more effective for ASR.

\paragraph{Spoken Language Modeling} 
Following the huge success of language models on text data \cite{devlin2018bert, radford2019gpt2, brown2020gpt3}, 
the Zero Resource Speech Challenge 2021 \cite{nguyen2020zero, dunbar2021zs21} opens up new possibilities
for learning high-level language properties from raw audio without any text labels.
They introduced 4 zero-shot evaluation metrics at different linguistic levels (phonetic, lexical, syntactic, semantic), along with composite baseline systems consisting of an acoustic discretization module (Contrastive Predictive Coding, or CPC+k-means) followed by a language model (BERT or LSTM) on the discretized units. The CPC model takes the raw audio as input and produces phonetic representations at a lower frame rate of 100Hz, helping the language model to learn high-level information from the raw audio.
In the same spirit, \cite{kushal2021gslm} introduced Generative Spoken Language Modeling (GSLM), the task of learning and generating spoken language from raw audio only. They provided baseline systems consisting of a discrete speech encoder (CPC, wav2vec 2.0, HuBERT), a generative language model (GPT-like model), and a speech decoder (Tacotron-2, \cite{shen2018tacotron}). The models are evaluated on spoken language modeling metrics \cite{nguyen2020zero}, ASR-based generation metrics \cite{kushal2021gslm} as well as human evaluation metrics.

\section{Experimental Setup}\label{section:metrics}
In this section, we first present the evaluation metrics as well as the dataset used to train and evaluate the models. We then explain our models and the inference methods for model evaluation.
\subsection{Evaluation Metrics}


We evaluate our models with the ZeroSpeech 2021 Benchmark Metrics \cite{nguyen2020zero}, consisting of 4 zero-shot tests probing for the quality of spoken language models at four linguistic levels: phonetic (Libri-light ABX metrics), lexical (sWUGGY spot-the-word metrics), syntactic (sBLIMP acceptability metrics) and semantic (sSIMI similarity metrics).

\paragraph{Libri-light ABX metrics} Given a pair of similar triphones (e.g., `aba'-`apa') spoken by a same speaker and an intervening sound (either `aba' or `apa'), the model has to tell which sound has a closer representation to the intervening sound. The ABX metrics is reported as the error rate that the model fails to choose the correct triphone.
\paragraph{sWUGGY spot-the-word metrics} Given a pair of a word and a similar non-word (e.g., `brick'-`blick'), the model has to tell which is the word based on their probability. The spot-the-word metrics is reported as the accuracy that the model assigns a higher probability to the word.
\paragraph{sBLIMP acceptability metrics} Given a linguistic minimal sentence pair of matched grammatical and ungrammatical sentences (e.g., `he loves it'-`he love it'), the model has to tell which is the grammatical sentence. The acceptability metrics is reported as the accuracy that the model assigns a higher probability to the grammatical sentence.
\paragraph{sSIMI similarity metrics} Given a pair of words (e.g., `happy'-`joyful'), the model has to compute a similarity score based on their representations.
The similarity metrics is reported as the Pearson correlation coefficient (PCC) between model scores and human judgements. In this work, the sSIMI scores are weighted across different subsets according to their sizes and averaged across LibriSpeech and synthetic subsets to make it more
accurate and consistent.
We reported it as wSIMI.

\subsection{Datasets}
\paragraph{Training Dataset}
We train our models on LibriSpeech \cite{Panayotov2015librispeech}, an English corpus containing 1000 hours of read speech based on public domain audio books.
The models are validated on LibriSpeech dev-clean and dev-other subsets, comprising 10 hours of speech in total.
\paragraph{Metrics Datasets}
The metrics datasets are either extracted sounds from LibriSpeech (ABX, sSIMI) or synthesised using Google API\footnote{\url{https://cloud.google.com/text-to-speech}} (sWUGGY, sBLIMP, sSIMI). The datasets containing words or sentences were filtered to only contain the LibriSpeech vocabulary (except sWUGGY non-words), and
are split into dev and test sets. The dev sets have been made publicly available at the ZeroSpeech 2021 Challenge website \footnote{\url{https://zerospeech.com/2021/instructions.html#evaluation-dataset}}.

\subsection{Models}
\paragraph{ZeroSpeech 2021 Baseline}
The ZeroSpeech 2021 Baseline System \cite{nguyen2020zero} is a composite of three components: an acoustic model (CPC, \cite{oord2018cpc,riviere2020unsupervised}), a clustering module (k-means) and a language model (BERT, \cite{devlin2018bert}). The CPC model is first trained to obtain good phonetic representations of the speech, which are then discretized into sequences of units with the k-means model. The BERT model is finally trained on these discrete units to better learn linguistic information.

As we only focus on the language modeling system in this work, we shall use the best CPC model in the ZeroSpeech 2021 Baseline System, which comprises a 5-layer 1D-CNN Encoder followed by a 4-layer LSTM autoregressive model.
The features are extracted from the 2nd layer (unless otherwise specified) of the LSTM model, with a rate of 100Hz, and are either discretized with a 50-unit k-means model (discrete) or left unchanged (continuous).

\paragraph{BERT with discrete and continuous features}
We modify the BERT model so that it is able to take as input either discrete units obtained from k-means or continuous features extracted from CPC, in which case the masking is done by replacing the features with a masked embedding vector.
We also allow the model to predict either discrete target units or continuous target features, with multiple choices of an appropriate objective for each case.
When predicting discrete targets, 
we use a cross-entropy objective (Negative Log-Likelihood, or NLL loss) but with two slightly different implementations.
We could simply employ a linear classification head at the output of the BERT model as usual (which we denote by \textit{linear NLL}, or NLL-l) or force the BERT output features to be similar to the embedding vectors of the target units as for HuBERT (cf. equation (3) from \cite{weining2021hubert}, we denote this by \textit{embedding NLL}, or NLL-e).
In the case of continuous targets,
it can be a reconstruction objective (L1 loss or L2 loss) or a contrastive objective (Noise Contrastive Estimation, or NCE loss). In the latter case, the predicted features are contrasted with 100 negative features sampled from the same phrase (similar to continuousBERT, \cite{baevski2020effectiveness}).

We use a BERT base model, which comprises a 12-layer Transformer Encoder. Our implementation is based on the wav2vec2.0 \cite{baevski2020w2v2} Transformer Encoder
\footnote{\url{https://github.com/pytorch/fairseq/tree/main/examples/wav2vec}} 
using fairseq \cite{myleott2019fairseq}. Each input sequence contains the features of a full audio file, and we consider at most 15.6 seconds of audio per file. We trained all models for 250k update steps on 32 GPUs, with a batch size of 175s per GPU. The learning rate was warmed up to a peak value of $1 \times 10 ^{-5}$ after 32k steps. For the masking, we
masked $M$ consecutive tokens for each span, where $M \sim \mathcal{N}(10, 10)$, with a total masking coverage of roughly half of the input tokens (spans may overlap).


\subsection{Model Inference for Evaluation}
\paragraph{ABX Distance}
For the ABX metrics, we extract frame-by-frame representation features for each audio file. Then, the ABX distance between two files is computed as the average angular distance 
of the representations along the realigned Dynamic Time Wrapping path. Given two audio files $x$ and $y$ with two sequences of representation $\textbf{r}^x = r^x_1,\dots,r^x_T$ and
$\textbf{r}^y = r^y_1,\dots,r^y_S$ respectively, the ABX distance between $x$ and $y$ is computed as follows: 
\begin{equation}\label{eq:ABX}
\begin{split}
    &d_{ABX}(x,y) =  \frac{1}{\lvert\text{path}_{\text{DTW}}(\textbf{r}^x,\textbf{r}^y)\rvert}\sum_{(i,j)\in\text{path}_{\text{DTW}}(\textbf{r}^x,\textbf{r}^y)} sim(r^x_i,r^y_j),
\end{split}
\end{equation}
where $sim(r^x_i,r^y_j)$ is the angular distance (in radian) between the embeddings $r^x_i$ and $r^y_j$.

We note that in this paper the ABX metrics are mainly used to evaluate the input and target features of the BERT model, and therefore the ABX distances are mostly performed on the CPC features without using the BERT model.

\paragraph{Probability Estimation}
For sWUGGY and sBLIMP metrics, 
we compute for each audio file a model-based pseudo log-probability (m-PLP) of the trained BERT model. 
Given an audio file $x$ with the input and target features for the BERT model $x_1...x_T$ and $\hat{x}_1...\hat{x}_T$ respectively, the m-PLP is computed as follows:
\begin{equation}\label{eq:span-PP}
\begin{split}
    & \text{m-PLP}(x) = 
    \quad \sum_{\substack{j = 0 \\ i = j \Delta t}}^{\lfloor (T-M)/\Delta t \rfloor} \sum_{m=1}^M PLP(\hat{x}_{i+m}|\overline{x_{i+1}..x_{i+M}}),
\end{split}
\end{equation}
where $M$ is a chosen size of a sliding window, $\Delta t$ is a chosen step of the sliding window 
and $PLP(\hat{x}_{i+m}|\overline{x_{i+1}..x_{i+M}})$ is a pseudo log-probability of the target $\hat{x}_{i+m}$ given by the BERT model with $M$-span masked inputs $x_1..x_im..mx_{i+M+1}..x_T$ ($m$ represents a masked feature).

For models with NLL or NCE loss, $PLP(\hat{x}_{i}|\overline{x_{i+1}..x_{i+M}})$ is computed as 
the log value of the probability given by the softmax layer of the BERT model (in the NLL case, the probability is computed over all tokens, while in the NCE case it is computed over all sampled negative examples).
For models with L1 or L2 loss, we compute $PLP(\hat{x}_{i}|\overline{x_{i+1}..x_{i+M}})$ as the negative reconstruction loss of the predicted feature and the target feature $\hat{x}_{i}$. 
The negativity ensures that a correct target has a higher m-PLP.

The m-PLP extends the span-masked pseudo probability (span-PP) \cite{nguyen2020zero} to BERT models with continuous targets. It is derived from the pseudo-loglikelihood score (PLL) for MLMs \cite{wang-cho-2019-bert}, which was shown to be an effective sentence scoring method for BERT models in many scenarios \cite{salazar-etal-2020-masked}.

The choice of $M$ and $\Delta t$ is determined for each model using the dev sets, and is given in Table \ref{tab:hyperparams}. In our experiments, we always consider $\Delta t = 5$ and vary $M$ in $\{15,25,35,45,55\}$. For models trained on HuBERT features (section \ref{sec:varyingdiscreteunits}), we vary $M$ in $\{5,10,15,20,25\}$ as the frame rate is 50Hz instead of 100Hz as for CPC.

\paragraph{Similarity Score}
For the sSIMI metrics,
we extract a fixed-length representation for each audio file by applying a pooling function (mean, max, min) over hidden features from one layer of the Transformer Encoder. 
The similarity score of two audio files is computed as the cosine similarity between the two corresponding representations.
The choice of the hidden layer and the pooling function is determined for each model using the dev sets and is given in Table \ref{tab:hyperparams}.

\section{Results} \label{section:results}
\subsection{Discrete bottleneck seems to be essential for spoken language modeling}
Table \ref{tab:discrete&continuous} reports the performances of our BERT models,
trained with either continuous or discrete CPC features of the LibriSpeech 960h dataset,
on lexical (sWUGGY), syntactic (sBLIMP) and semantic (wSIMI) metrics.

\begin{table}[ht]

\begin{center}
\begin{adjustbox}{max width=0.5\textwidth, center}
\begin{tabular}{c @{\hspace{0.7\tabcolsep}} l @{\hspace{0.9\tabcolsep}}l@{\hspace{0.9\tabcolsep}} l @{\hspace{0.9\tabcolsep}} |c@{\hspace{0.9\tabcolsep}} c@{\hspace{0.9\tabcolsep}}c}
\textbf{id} & \textbf{input} & \textbf{target} &\textbf{loss} & sWUGGY↑& sBLIMP↑ & wSIMI↑ \\

\midrule
\midrule

& \multicolumn{6}{l}{\rule{0pt}{2ex}\textit{discrete input, discrete target}} \\
1 & \rule{0pt}{2.5ex}disc. & disc. & NLL-l & 79.28 & 59.71 & 6.32 \\
2 & disc. & disc. & NLL-e & \underline{\textbf{80.02}} & \underline{\textbf{59.86}} & \textbf{7.87} \\
\midrule
& \multicolumn{6}{l}{\rule{0pt}{2ex}\textit{continuous input, discrete target}} \\
3 & \rule{0pt}{2.5ex}\textit{cont.} & disc. & NLL-l & \textbf{60.36} & \textbf{53.23} & 8.39 \\
4 & \textit{cont.} & disc. & NLL-e & 60.20 & 52.78 & \textbf{9.49} \\
\midrule
& \multicolumn{6}{l}{\rule{0pt}{2ex}\textit{continuous input, continuous target}} \\
5 & \rule{0pt}{2.5ex}\textit{cont.} & \textit{cont.} & NCE & 56.84 & 52.62 & \textbf{9.16} \\
6 & \textit{cont.} & \textit{cont.} & L1 & 59.23 & 53.12 & 7.85 \\
7 & \textit{cont.} & \textit{cont.} & L2 & \textbf{60.56} & \textbf{53.33} & 6.55 \\
\midrule
& \multicolumn{6}{l}{\rule{0pt}{2ex}\textit{discrete input, continuous target}} \\
8 & \rule{0pt}{2.5ex}disc. & \textit{cont.} & NCE & 65.69 & \textbf{57.24} & 9.33 \\
9 & disc. & \textit{cont.} & L1 & 73.93 & 56.02 & \underline{\textbf{10.69}} \\
10 & disc. & \textit{cont.} & L2 & \textbf{74.22} & 55.75 & 5.97 \\

\end{tabular}
\end{adjustbox}
\end{center}
\caption{
Performances on the dev sets of sWUGGY, sBLIMP, wSIMI metrics of BERT models using either continuous ZeroSpeech CPC features (layer 2 of the LSTM module of CPC-big) or discretized features (with a 50-unit k-means model) as inputs and targets.
Best scores in each category are in bold, best scores overall are underlined.
}
\label{tab:discrete&continuous}
\end{table}

We first examine how the continuity of the input and target features affects the quality of the BERT model on the evaluation metrics.
By comparing the best scores in each case, we see that 
having discrete inputs helps the model learn better lexical and syntactic information, whereas models with continuous inputs do have better than chance performance on the lexical task.
We observe that the best models on the language model tasks are obtained with discrete inputs and discrete targets, which is the classic configuration of BERT.
Predicting continuous targets from discrete inputs, where the model acts as an autoencoder decoder, is also beneficial and nearly catches up with the best models.
It is interesting, still, to note that it is possible to acquire some language information without any discretization.
The wSIMI scores are still quite low, but we see in general that having continuous information does help.



\subsection{Is continuous input always bad?}
We observe during our training experiments that the masked prediction objective is too easy for some models with continuous inputs and could quickly lead to overfitting. This could be explained by the fact that the input and target features are extracted from the same layer of the LSTM autoregressive module of CPC.
As a consequence, we try using the input features from different layers of the LSTM module, while maintaining the same target layer. We keep using the NLL-e loss for discrete targets while using NCE and L1 loss for continuous targets. The results are reported in Table \ref{tab:multipleinputviews}.

\begin{table}[ht]

\begin{center}
\begin{adjustbox}{max width=0.5\textwidth, center}
\begin{tabular}{ll|cclcc}
& & \multicolumn{2}{c}{ABX within↓} && \multicolumn{2}{c}{ABX across↓}  \\
& & clean  & other   && clean & other \\
\midrule
\midrule
\multirow{2}{*}{layer 0} & \textit{cont.} & 11.50 & 14.09 && 18.53 & 24.70 \\
& disc. & 21.46 & 24.21 && 30.77 & 34.91 \\
\midrule
\multirow{2}{*}{layer 2} & \textit{cont.} & 3.41 & 4.84 && 4.20 & 7.65 \\
& disc. & 6.38 & 10.22 && 8.22 & 14.86 \\
\midrule
\multirow{2}{*}{layer 4} & \textit{cont.} & 9.49 & 11.95 && 10.01 & 15.70 \\
& disc. & 19.81 & 21.64 && 24.39 & 28.04 \\
\end{tabular}
\end{adjustbox}
\end{center}
\caption{Within and Across Speaker ABX error (lower is better) on Libri-light dev-clean and -other for continuous and discretized features of different layers of the LSTM autoregressive module of CPC-big model. Layer 0 means the output of the CNN Encoder module.}
\label{tab:ABX}
\end{table}

We observe that using continuous input features from a different layer does reduce overfitting during training, which significantly improves the performances of the models on LM metrics, especially for sWUGGY scores. Interestingly, we note that using continuous input features from a lower LSTM layer (layer 0, where the ABX errors are high, cf. Table \ref{tab:ABX}) to predict target features from a higher LSTM layer (layer 2) is more beneficial to the model than using high quality continuous input features from the same or higher layer as the target features (layer 2, layer 4). This is not the case, however, for discrete input models, where the model benefits from good quality input units. 

\begin{table}[ht]

\begin{center}
\begin{adjustbox}{max width=0.5\textwidth, center}
\begin{tabular}{c @{\hspace{0.7\tabcolsep}} l @{\hspace{0.9\tabcolsep}}l @{\hspace{0.9\tabcolsep}}| c@{\hspace{0.9\tabcolsep}} c@{\hspace{0.9\tabcolsep}}c}
\textbf{id} & \rule{0pt}{2ex}\textbf{input} & \textbf{target} & sWUGGY↑ & sBLIMP↑ & wSIMI↑ \\
\midrule
\midrule
\multicolumn{6}{l}{\rule{0pt}{2ex}
\quad \textit{discrete input, discrete target, NLL-e loss}
} \\
2 & layer 2 &layer 2 & \underline{\textbf{80.02}} & \underline{\textbf{59.86}} & 7.87 \\
11 & \rule{0pt}{2ex}layer 0 &layer 2 & 64.91 & 52.45 & 7.48 \\
12 & layer 4 &layer 2 & 70.68 & 55.06 & \textbf{8.61} \\
\midrule
\multicolumn{6}{l}{\rule{0pt}{2ex}
\quad \textit{continuous input, discrete target, NLL-e loss}
} \\
4 & \textit{layer 2} &layer 2 & 60.20 & 52.78 & \textbf{9.49} \\
13 & \rule{0pt}{2ex}\textit{layer 0} &layer 2 & \textbf{77.19} & \textbf{55.30} & 7.25 \\
14 & \textit{layer 4} &layer 2 & 67.41 & 54.13 & 8.06 \\
\midrule
\multicolumn{6}{l}{\rule{0pt}{2ex}
\quad \textit{continuous input, continuous target, NCE loss}
} \\
5 & \textit{layer 2} & \textit{layer 2} & 56.84 & 52.62 & \textbf{9.16} \\
15 & \rule{0pt}{2ex}\textit{layer 0} & \textit{layer 2} & \textbf{65.53} & \textbf{55.20} & 6.17 \\
16 & \textit{layer 4} & \textit{layer 2} & 59.81 & 52.93 & 8.32 \\
\midrule
\multicolumn{6}{l}{\rule{0pt}{2ex}
\quad \textit{continuous input, continuous target, L1 loss}
} \\
6 & \textit{layer 2} & \textit{layer 2} & 59.23 & 53.12 & \textbf{7.85} \\
17 & \rule{0pt}{2ex}\textit{layer 0} & \textit{layer 2} & \textbf{67.83} &\textbf{53.59} & 7.25 \\
18 & \textit{layer 4} & \textit{layer 2} & 63.68 & 53.26 & 6.90 \\
\midrule
\multicolumn{6}{l}{\rule{0pt}{2ex}
\quad \textit{discrete input, continuous target, NCE loss}
} \\
8 & layer 2 & \textit{layer 2} & \textbf{65.69} & \textbf{57.24} & \textbf{9.33} \\
19 & \rule{0pt}{2ex}layer 0 & \textit{layer 2} & 58.55 & 52.31 & 8.07 \\
20 & layer 4 & \textit{layer 2} & 58.61 & 54.48 & 7.72 \\
\midrule
\multicolumn{6}{l}{\rule{0pt}{2ex}
\quad \textit{discrete input, continuous target, L1 loss}
} \\
9 & layer 2 & \textit{layer 2} & \textbf{73.93} & \textbf{56.02} & \underline{\textbf{10.69}} \\
21 & \rule{0pt}{2ex}layer 0 & \textit{layer 2} & 62.98 & 53.47 & 5.20 \\
22 & layer 4 & \textit{layer 2} & 65.92 & 53.94 & 7.01 \\

\end{tabular}
\end{adjustbox}
\end{center}
\caption{
Performances on the dev sets of sWUGGY, sBLIMP, wSIMI metrics of BERT models using the input features from different layers of the LSTM module of CPC-big model.
Layer 0 means the output of the CNN Encoder module.
Best scores in each category are in bold, best scores overall are underlined.
}
\label{tab:multipleinputviews}
\end{table}


Overall, we observe that 
discrete-discrete model (with the same input and target units) yields the best performance when the quality of discrete units is good.
Continuous-discrete is also a great choice when using low-level input features. When there are no discrete units at all, the LM performances are still limited, even if using a NCE loss could help a bit with syntactic and semantic metrics. 

\subsection{Varying the number of discrete units} \label{sec:varyingdiscreteunits}

Here, we address the question as to why discrete units are better than continuous ones. One hypothesis is that discrete units manage to remove linguistically irrelevant information and force the transformer to focus on linguistic ones. To test this, we run a speaker discrimination probe on the discrete units and continuous features. In addition, we run a new experiment varying the number of discrete units from 20 to 2000. Hypothetically, when the number of units is too small (eg, smaller than the number of phonemes), the resulting phonetic confusions should degrade the learning of higher linguistic representations. Conversely, when the number of units is too large, the quantization step would start to leak other-than-phonetic information into the representation, hence making it closer to the continuous representations. 
\begin{table}[ht]

\begin{center}
\begin{adjustbox}{max width=0.5\textwidth, center}
\begin{tabular}{c@{\hspace{0.5\tabcolsep}}l@{\hspace{0.5\tabcolsep}}|c@{\hspace{0.9\tabcolsep}}c|c@{\hspace{0.3\tabcolsep}}c@{\hspace{0.9\tabcolsep}}c}
& & \multicolumn{2}{c|}{\textbf{unit quality}} & \multicolumn{3}{c}{\textbf{language modeling on units}} \\
\textbf{model} & \textbf{n units} & spk prb↑ & ABX↓ &  sWUGGY↑ & sBLIMP↑ & sSIMI↑ \\
\midrule
\midrule
\multirow{7}{*}{CPC} & 20 & 30.40 & 12.66 & 71.71 & 58.97 & 4.72 \\
& 50 & 34.00 & 9.89 & 80.02 & \textbf{59.86} & \textbf{7.87} \\
& 100 & 49.20 & 9.56 & \textbf{80.47} & 59.47 & 6.09 \\
& 200 & 56.00 & 9.72 & 79.90 & 58.90 & 4.45 \\
& 500 & 64.00 & 10.72 & 79.66 & 59.72 & 6.37 \\
& 1000 & 61.60 & 11.99 & 79.86 & 58.46 & 6.78 \\
& 2000 & 67.60 & 14.24 & 78.46 & 58.25 & 5.94 \\
& \textit{cont.} & \textit{98.00} &  \textit{5.02} & \multicolumn{3}{c}{\textit{-}} \\
\midrule
\multirow{7}{*}{HuBERT} & 20 & 24.40 & 14.04 & 62.89 & 57.06 & 7.80 \\
& 50 & 38.00 & 9.19 & 76.79 & 61.12 & 8.61 \\
& 100 & 48.00 & 8.34 & 81.09 & 62.47 & 5.19 \\
& 200 & 61.60 & 7.57 & 81.54 & 62.78 & 7.03 \\
& 500 & 68.40 & 7.73 & \textbf{83.06} & \textbf{62.89} & 9.73 \\
& 1000 & 74.40 & 9.04 & 82.58 & 61.55 & 8.64 \\
& 2000 & 73.20 & 11.00 & 81.61 & 62.85 & \textbf{10.66} \\
& \textit{cont.} & \textit{99.60} &  \textit{4.23} & \multicolumn{3}{c}{\textit{-}} \\
\midrule
\textit{Forced Phones} & \textit{40} & \textit{10.00} & \textit{0.00} & \textit{92.19} & \textit{63.72} & \textit{6.23}

\end{tabular}
\end{adjustbox}
\end{center}
\caption{Discrete Unit quality (Speaker probing and ABX) and Performance of the BERT models trained on Discrete Units on the dev sets of LM scores (sWUGGY, sBLIMP, wSIMI) for different numbers of clusters on CPC and HuBERT features. The ABX is averaged on dev-clean and dev-other within and across subsets.}
\label{tab:nbcluster}
\end{table}

To support our hypothesis, we run kmeans on
the continuous features of both CPC and HuBERT models,
and vary k to be 20, 50, 100, 200, 500, 1000, and 2000, after which we train a discrete-discrete BERT model. 
For the CPC features, we take the layer 2 features of the CPC-big model as usual. For the HuBERT features, we train our own HuBERT base model as described in Section \ref{sec:overall_test_results}, we then take the features from layer 12 of the Transformer Encoder after the 2nd iteration, which have the best ABX (cf. Table \ref{tab:BERTlayersABX}).
Following \cite{kharitonov2022textlesslib}, we train a speaker classifier in the following way: We randomly split LibriSpeech dev-clean utterances into train/valid/test (80\%/10\%/10\%) sets and train a two-layer Transformer classifier on the sequences of discrete units or continuous features of the utterances.
The classification head is performed on the first token (\texttt{bos}, or begin-of-sentence) of the transformer outputs.
For this speaker probing task, there are 40 classes (speakers).
The models are trained for 20 epochs and are validated on the valid set. We finally report the test accuracy.
For reference, we also include the forced phonemes units (frame-by-frame phonemes). As the forced phonemes contain a silence, there are 40 units in total.

\begin{figure}[t]
     \centering
     \includegraphics[width=0.4\textwidth]{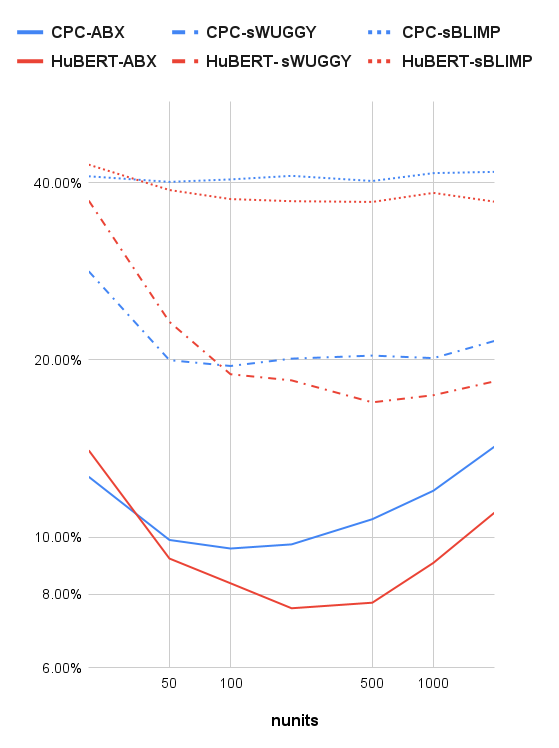}
     \caption{ABX of Discrete units and Error rate on the dev sets of LM scores (sWUGGY, sBLIMP) for different numbers of clusters for CPC and HuBERT features. The ABX is averaged on dev-clean and dev-other within and across subsets.}
     \label{fig:nbcluster}
\end{figure}

The results are reported in Table  \ref{tab:nbcluster} and illustrated in Figure \ref{fig:nbcluster}. As expected speaker classification accuracy increases with the number of clusters, and the continuous features yield the best classification. We can observe a U-shaped curve in performance across the different language metrics as a function of the number of units. Interestingly, the optimum number of units seems to be different across the model features (CPC, HuBERT) and linguistic levels.
HuBERT features are better than CPC features in most  cases, and seem to benefit from more clusters than CPC features.
In general, we see that the language model scores seem to decrease slowly compared to ABX as the number of clusters becomes bigger.
It is also interesting to note that the language model scores become steadily good as soon as the number of clusters is higher than the number of phonemes (40 units). This could be seen in Figure \ref{fig:phoneunitalign}, where we observe a limited unit-phoneme correspondence when having only 20 discrete units; but as soon as the number of clusters reaches 50, we see a clear correspondence between the units and the phonemes, although several "hard" phonemes are still dispersed and don't correspond to a single unit (e.g.
ch, oy, th, uh); 
when there are 500 clusters, 
there are more units representing a single phoneme, and 
most "hard" phonemes are now assigned by certain units.

These results support the hypothesis that the superiority of the discrete units is due to the fact that they block the propagation and amplification of non-linguistic signals that may be present (even if attenuated) in continuous representations.

\subsection{Comparison with state-of-the-art systems}\label{sec:overall_test_results}
\begin{table*}[ht]
\begin{center}
\begin{adjustbox}{max width=1.0\textwidth, center}
\begin{tabular}{c@{\hspace{0.7\tabcolsep}} lll | c@{\hspace{0.5\tabcolsep}}c c@{\hspace{0.5\tabcolsep}}c| ccc}
&&&& \multicolumn{4}{c|}{ABX (target features)↓} &&&\\
& \multicolumn{3}{l|}{\textit{Systems}} & \multicolumn{2}{c}{within} & \multicolumn{2}{c|}{across} & \multirow{2}{*}{sWUGGY↑} & \multirow{2}{*}{sBLIMP↑} & \multirow{2}{*}{wSIMI↑} \\
\textbf{id} & \textbf{input} & \textbf{target} & \textbf{loss} & clean & other & clean & other & & & \\

\midrule
\midrule
& \multicolumn{10}{l}{\rule{0pt}{2ex} \textit{ZeroSpeech 2021 Best Baseline System} \cite{nguyen2020zero}} \\
&CPC-layer2+km50 &CPC-layer2+km50 & NLL-l & 6.71 &	10.62 &	8.41 & 15.06 &	75.51 &	56.16 &	2.05\\
\midrule
& \multicolumn{10}{l}{\rule{0pt}{2ex} \textit{ZeroSpeech 2021 Text Topline Systems} \cite{nguyen2020zero}} \\
& Forced phones & Forced phones & NLL-l & 0.00&	0.00&	0.00&	0.00&	91.88&	63.16 &	4.44\\
& Phones & Phones & NLL-l & - & -&	-&	-&	97.67&	66.91&	12.80\\
\midrule
\midrule
& \multicolumn{10}{l}{\rule{0pt}{2ex} \textit{BERT Models on CPC-big Features}} \\
2 &CPC-layer2+km50 &CPC-layer2+km50 & NLL-e & 6.71 &	10.62 &	8.41 & 15.06 & \textbf{80.29} & \textbf{59.93} & 6.56 \\
13 & \textit{CPC-layer0} & CPC-layer2+km50 & NLL-e & 6.71 &	10.62 &	8.41 & 15.06 & 77.22 & 55.62 & \textbf{6.61} \\
17 & \textit{CPC-layer0} & \textit{CPC-layer2} & L1 & 3.28 &	4.81 &	4.31 & 7.92 & 68.37 & 53.95 & 5.68 \\
9 &CPC-layer2+km50 & \textit{CPC-layer2} & L1 & 3.28 &	4.81 &	4.31 & 7.92 & 74.46 & 55.38 & 6.17 \\
\midrule
& \multicolumn{10}{l}{\rule{0pt}{2ex} \textit{HuBERT Base Models}} \\
23 & \textit{waveform} & MFCC+km100 & NLL-e & 20.22 & 24.97 &	33.42 & 40.45 & 62.74 & 54.11 & 5.58 \\
24 & \textit{waveform} & H-iter1-layer6+km500 & NLL-e & 6.29 & 7.51 &	8.76 & 12.82 & 79.13 & 58.89 & 5.45 \\
25 & \textit{waveform} & H-iter2-layer12+km500 & NLL-e & 5.87 & 7.15 &	6.96 & 10.73 & \textbf{80.19} & \textbf{59.29} & \textbf{5.87} \\
\midrule
& \multicolumn{10}{l}{\rule{0pt}{2ex} \textit{BERT Models on HuBERT Discrete Units}} \\
26 & H-iter2-layer12+km500 & H-iter2-layer12+km500 & NLL-e & 5.87 & 7.15 &	6.96 & 10.73 & \underline{\textbf{83.29}} & \underline{\textbf{61.93}} & \underline{\textbf{9.73}}

\end{tabular}
\end{adjustbox}
\end{center}
\caption{
Comparison on the test sets of the 4 ZeroSpeech 2021 metrics of our 
BERT models trained on continuous or discrete CPC features,
BERT model trained on HuBERT discrete units and
HuBERT Base models with ZeroSpeech 2021 Baseline and Topline Systems. 
For each continuous/discrete combination, we choose the best performing model on the dev set as reported in Table \ref{tab:multipleinputviews}.
We trained the HuBERT model for 3 iterations.
The targets used to train the 3 iterations are discretized MFCC features (100 units), discretized features from Transformer's layer6 of 1st iteration (500 units) and discretized features from Transformer's layer12 of 2nd iteration (500 units) respectively.
All models were trained on the LibriSpeech 960h dataset.
For the ABX metrics, we report the scores on the target features used to train the model. Best scores in each category are in bold, best scores overall are underlined.
}
\label{tab:overall}
\end{table*}
We evaluate the HuBERT model on the zero-shots metrics and compare the results with our trained BERT models.
The HuBERT model is trained iteratively, using 
clustering units from
features of previous iteration as the teacher.
We trained a HuBERT base model, which comprises a 7-layer CNN Encoder followed by a 12-layer Transformer Encoder, on the Librispeech 960h dataset for 3 iterations. The teachers for each iteration are
MFCC features (100 units), Transformer's layer6 of 1st iteration (500 units) and Transformer's layer12 of 2nd iteration (500 units) respectively.
Architecturally, the Transformer Encoder of the HuBERT model 
is very similar to our model 13 (\textit{continuous input layer 0, discrete target layer 2, NLL-e loss})
where they both take as input the continuous features of the CNN Encoder and predict discrete targets obtained from features of a higher level with a NLL-eloss. 

Overall performances on the ZeroSpeech 2021 test sets are reported in Table \ref{tab:overall}.
For each of discrete/continuous combinations,  we choose the best performing model on the dev set as reported in Table \ref{tab:multipleinputviews}. We 
also include the discrete-discrete model trained on HuBERT Discrete Units (500 units), which was reported to have the best LM scores in section \ref{sec:varyingdiscreteunits}.
We first observe a huge improvement of model 2 compared with the baseline system, even if they both use the same units for the BERT model. This improvement greatly comes from the reimplementation of the BERT model, which uses the wav2vec2 Transformer Encoder model\footnote{One main difference between the two Transformer models is that wav2vec2 uses a Convolutional Positional Embedding instead of the standard Sinusoidal Positional Embedding. However, we did not study the effect of this difference in this paper.}. 
Changing the NLL-l loss to NLL-e loss also improves a little bit (cf. Table \ref{tab:discrete&continuous}).

It seems that using good quality discrete units as targets is very beneficial for the language models, achieving better scores than using continuous targets in all the metrics.
The HuBERT model performs surprisingly well, approaching our best model on the language model tasks. This means that the Transformer Encoder of HuBERT acts as a language model as well. We see that as soon as the discrete targets have better quality, the HuBERT model manages to have better results on spoken language modeling metrics. We see that the discrete-discrete model on HuBERT Discrete Units (model 26) further improves the scores on all the metrics, confirming again our finding that it's better to train a discrete-discrete model when we have good quality units.

Comparing the results with the ZeroSpeech 2021 Systems, we observe that our models are closing the gap between spoken and text-based language models. 


\section{Conclusion}
This work analyses the 
importance of discretization in spoken language modeling. 
We experimentally show that discretization is essential for spoken language modeling, 
although high-quality discrete units are required to obtain good performances.
We also show the possibility of learning high-level language properties of a self-supervised speech representation learning model like HuBERT. 
Finally, we obtain state-of-the-art results on 3 out of 4 metrics of the Zero Resource Speech Challenge 2021 (Track 1 - Speech Only), bridging the gap between speech and text-based systems.
Note though that because HuBERT requires a teacher that learns a discrete representation, the overall training of HuBERT is not end-to-end, because the training of the teacher is not (in fact, requires several iterations). Further work is needed to simplify this kind of training loop to learn language directly from speech inputs. Further work is also needed to assess whether the present results can generalize to other languages and datasets.  

\section*{Acknowledgments}


In this work, ED in his EHESS role was supported by the Agence Nationale pour la Recherche (ANR-17-EURE-0017 Frontcog, ANR-10-IDEX-0001-02 PSL*, ANR-19-P3IA-0001 PRAIRIE 3IA Institute), a grant from CIFAR (Learning in Machines and Brains) and HPC resources from GENCI–IDRIS (Grant 2021-AD011011691R1).

We would like to thank the reviewers for their thoughtful comments on the paper.
We also thank
Jade Copet, Evgeny Kharitonov, Morgane Riviere, Paden Tomasello, Wei-Ning Hsu, Yossef Mordechay Adi, Abdelrahman Mohamed,
Maureen de Seyssel, Marvin Lavechin, Robin Algayres, Xuan-Nga Cao, Nicolas Hamilakis, Hadrien Titeux, Gwendal Virlet, Marianne Metais
for helpful discussions on the paper.

\bibliographystyle{IEEEtran}
\bibliography{custom}

\appendix
\section{Appendix}
\subsection{Hyper-parameters of trained models}\label{section:hyperparams}
Table \ref{tab:hyperparams} shows the hyperparameters of all trained models in the paper.

\subsection{Unit-Phone alignments for CPC Discrete Units}
\begin{figure}[ht]
     \centering
     \includegraphics[width=0.5\textwidth]{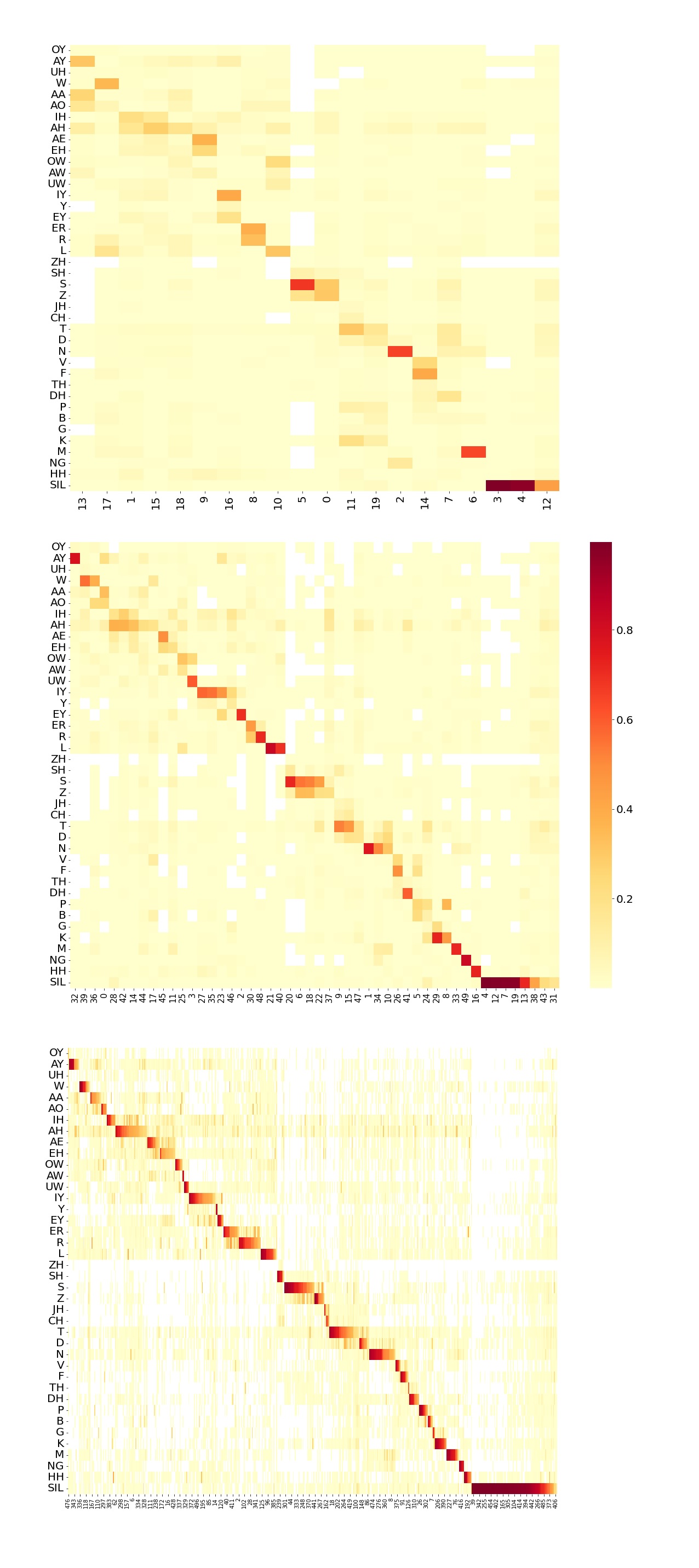}
     \caption{Probability that each discrete unit belongs to possible phonemes $P\left(phoneme\mid unit\right)$ for discrete units obtained by clustering CPC features with different numbers of clusters: 20 (top), 50 (middle) and 500 (bottom). Unit-Phoneme alignments are collected on Librispeech dev-clean subset. The phoneme order is obtained by clustering the rows of the 50-unit model with a hierarchical clustering method.}
     \label{fig:phoneunitalign}
\end{figure}

We analyse to what extent the discrete units obtained with different numbers of clusters correlate with the gold phonemes.  
Using the phoneme alignments of Librispeech available from \cite{nguyen2020zero}, we collect all unit-phoneme pairs from the utterances of the dev-clean subset and compute the probability of each phoneme given a discrete unit. Figure \ref{fig:phoneunitalign} (top, middle, bottom) shows this unit-phone alignment for discrete units obtained from CPC features with 20, 50 and 500 clusters respectively.
The phoneme order is obtained by clustering the rows of the 50-unit model with a hierarchical clustering method.

\subsection{Layer-wise analysis of BERT models on ABX metrics}
In addition to performing ABX on input and target features of the BERT models, we also compute the ABX error on the features extracted from hidden Transformer layers of the trained BERT/HuBERT models. Table \ref{tab:BERTlayersABX} reports the average (within-across) ABX errors on the Librispeech dev-clean subset for all the trained models in the paper.

We see that well-trained models with good language modeling scores (models 1,2,13,15,17,8,9,10; cf. Table \ref{tab:multipleinputviews}) seem to have very good ABX errors compared to the others. By looking at the best hidden features of each model, we see that models with discrete targets (models 1,2,13) are able to reconstruct hidden features which are better than the targets, while this is not the case for most models with continuous targets (except model 10).


\begin{table*}[h]

\begin{center}
\begin{adjustbox}{max width=0.7\textwidth, center}
\begin{tabular}{c @{\hspace{0.7\tabcolsep}} cccccc | ccllll}
& \multicolumn{6}{c|}{Training hyperparameters} & \multicolumn{6}{c}{Inference hyperparameters} \\
&  \multicolumn{2}{c}{n units (if discrete)}  & feat. & \multicolumn{2}{c}{masking} & num & \multicolumn{2}{c}{prob. est.} & \multicolumn{2}{c}{SIMI librispeech} & \multicolumn{2}{c}{SIMI synthetic} \\
id & input & target & stride & length & prob. & updates & $M$ & $\Delta t$ & layer & pooling & layer & pooling \\

\midrule
\midrule

& \multicolumn{9}{l}{\rule{0pt}{2ex} \textit{BERT Models on CPC-big Features}} \\

1 & 50 & 50 & 10ms & 10 & 0.5 & 250k & 15 & 5 & 11 & max & 1 & min \\
2 & 50 & 50 & 10ms & 10 & 0.5 & 250k & 15 & 5 & 5 & min & 7 & mean \\
11 & 50 & 50 & 10ms & 10 & 0.5 & 250k & 15 & 5 & 11 & mean & 10 & min \\
12 & 50 & 50 & 10ms & 10 & 0.5 & 250k & 15 & 5 & 9 & mean & 1 & min \\
\midrule

3 & - & 50 & 10ms & 10 & 0.5 & 250k & 35 & 5 & 6 & min & 4 & max \\
4 & - & 50 & 10ms & 10 & 0.5 & 250k & 35 & 5 & 8 & mean & 11 & min \\
13 & - & 50 & 10ms & 10 & 0.5 & 250k & 25 & 5 & 10 & max & 10 & min \\
14 & - & 50 & 10ms & 10 & 0.5 & 250k & 35 & 5 & 4 & max & 6 & mean \\
\midrule

5 & - & - & 10ms & 10 & 0.5 & 250k & 45 & 5 & 8 & max & 1 & mean \\
15 & - & - & 10ms & 10 & 0.5 & 250k & 35 & 5 & 11 & max & 11 & mean \\
16 & - & - & 10ms & 10 & 0.5 & 250k & 45 & 5 & 12 & mean & 12 & mean \\

6 & - & - & 10ms & 10 & 0.5 & 250k & 35 & 5 & 1 & mean & 12 & mean \\
17 & - & - & 10ms & 10 & 0.5 & 250k & 25 & 5 & 12 & min & 12 & max \\
18 & - & - & 10ms & 10 & 0.5 & 250k & 35 & 5 & 1 & mean & 12 & mean \\
7 & - & - & 10ms & 10 & 0.5 & 250k & 35 & 5 & 3 & mean & 1 & mean \\
\midrule

8 & 50 & - & 10ms & 10 & 0.5 & 250k & 25 & 5 & 12 & mean & 5 & min \\
19 & 50 & - & 10ms & 10 & 0.5 & 250k & 25 & 5 & 8 & mean & 4 & min \\
20 & 50 & - & 10ms & 10 & 0.5 & 250k & 25 & 5 & 8 & mean & 11 & mean \\

9 & 50 & - & 10ms & 10 & 0.5 & 250k & 15 & 5 & 12 & max & 12 & min \\
21 & 50 & - & 10ms & 10 & 0.5 & 250k & 15 & 5 & 11 & mean & 7 & min \\
22 & 50 & - & 10ms & 10 & 0.5 & 250k & 15 & 5 & 4 & mean & 12 & max \\
10 & 50 & - & 10ms & 10 & 0.5 & 250k & 15 & 5 & 10 & max & 12 & max \\
\midrule
& \multicolumn{9}{l}{\rule{0pt}{2ex} \textit{HuBERT Base Models}} \\
23 & - & 100 & 20ms & 10 & 0.65 & 250k & 15 & 5 & 2 & max & 5 & min \\
24 & - & 500 & 20ms & 10 & 0.65 & 400k & 15 & 5 & 1 & mean & 9 & max \\
25 & - & 500 & 20ms & 10 & 0.65 & 400k & 15 & 5 & 10 & mean & 6 & max \\
\midrule
& \multicolumn{9}{l}{\rule{0pt}{2ex} \textit{BERT Models on HuBERT Discrete Units}} \\
26 & - & 500 & 20ms & 10 & 0.5 & 250k & 10 & 5 & 7 & mean & 8 & mean \\
\end{tabular}
\end{adjustbox}
\end{center}
\caption{
Hyperparameters of all trained models.
}
\label{tab:hyperparams}
\end{table*}

\begin{table*}[h]
\begin{center}
\begin{adjustbox}{max width=0.7\textwidth, center}
\begin{tabular}{c@{\hspace{0.7\tabcolsep}} lll | c|cccc|c}

\textbf{id} & \textbf{input feature} & \textbf{target feature} & \textbf{loss} & input & layer 3 & layer 6 & layer 9 & layer 12 & target \\

\midrule
\midrule
& \multicolumn{9}{l}{\rule{0pt}{2ex} \textit{BERT Models on CPC-big Features}} \\
1 & CPC-l2+km50 & CPC-l2+km50 & NLL-l & \ABXbox{7.3} & \ABXbox{5.90} & \ABXbox{6.13} & \ABXbox{5.64} & \ABXbox{3.87} & \ABXbox{7.3}\\
2 & CPC-l2+km50 & CPC-l2+km50 & NLL-e & \ABXbox{7.3} & \ABXbox{5.36} & \ABXbox{6.97} & \ABXbox{5.40} & \ABXbox{3.95} & \ABXbox{7.3}\\
11 & CPC-l0+km50 & CPC-l2+km50 & NLL-e & \ABXbox{26.11} & \ABXbox{15.08} & \ABXbox{12.16} & \ABXbox{10.63} & \ABXbox{7.01} & \ABXbox{7.3}\\
12 & CPC-l4+km50 & CPC-l2+km50 & NLL-e & \ABXbox{22.1} & \ABXbox{13.57} & \ABXbox{9.40} & \ABXbox{10.05} & \ABXbox{6.06} & \ABXbox{7.3}\\
\midrule
3 & \textit{CPC-l2} & CPC-l2+km50 & NLL-l & \ABXbox{3.81} & \ABXbox{8.87} & \ABXbox{15.30} & \ABXbox{14.91} & \ABXbox{9.10} & \ABXbox{7.3}\\
4 & \textit{CPC-l2} & CPC-l2+km50 & NLL-e & \ABXbox{3.81} & \ABXbox{8.47} & \ABXbox{16.82} & \ABXbox{21.47} & \ABXbox{12.84} & \ABXbox{7.3}\\
13 & \textit{CPC-l0} & CPC-l2+km50 & NLL-e & \ABXbox{15.01} & \ABXbox{6.57} & \ABXbox{4.74} & \ABXbox{4.04} & \ABXbox{4.07} & \ABXbox{7.3}\\
14 & \textit{CPC-l4} & CPC-l2+km50 & NLL-e & \ABXbox{9.75} & \ABXbox{7.03} & \ABXbox{8.78} & \ABXbox{10.49} & \ABXbox{4.98} & \ABXbox{7.3}\\
\midrule
5 & \textit{CPC-l2} & \textit{CPC-l2} & NCE & \ABXbox{3.81} & \ABXbox{6.47} & \ABXbox{6.83} & \ABXbox{6.09} & \ABXbox{5.98} & \ABXbox{3.81}\\
15 & \textit{CPC-l0} & \textit{CPC-l2} & NCE & \ABXbox{15.01} & \ABXbox{7.45} & \ABXbox{5.63} & \ABXbox{4.73} & \ABXbox{5.93} & \ABXbox{3.81}\\
16 & \textit{CPC-l4} & \textit{CPC-l2} & NCE & \ABXbox{9.75} & \ABXbox{6.82} & \ABXbox{7.19} & \ABXbox{5.46} & \ABXbox{5.50} & \ABXbox{3.81}\\
6 & \textit{CPC-l2} & \textit{CPC-l2} & L1 & \ABXbox{3.81} & \ABXbox{7.41} & \ABXbox{7.81} & \ABXbox{10.31} & \ABXbox{12.87} & \ABXbox{3.81}\\
17 & \textit{CPC-l0} & \textit{CPC-l2} & L1 & \ABXbox{15.01} & \ABXbox{6.64} & \ABXbox{4.73} & \ABXbox{4.71} & \ABXbox{5.48} & \ABXbox{3.81}\\
18 & \textit{CPC-l4} & \textit{CPC-l2} & L1 & \ABXbox{9.75} & \ABXbox{5.89} & \ABXbox{5.27} & \ABXbox{4.80} & \ABXbox{5.38} & \ABXbox{3.81}\\
7 & \textit{CPC-l2} & \textit{CPC-l2} & L2 & \ABXbox{3.81} & \ABXbox{6.35} & \ABXbox{6.58} & \ABXbox{7.65} & \ABXbox{9.20} & \ABXbox{3.81}\\
\midrule
8 & CPC-l2+km50 & \textit{CPC-l2} & NCE & \ABXbox{7.3} & \ABXbox{6.49} & \ABXbox{5.50} & \ABXbox{6.87} & \ABXbox{5.37} & \ABXbox{3.81}\\
19 & CPC-l0+km50 & \textit{CPC-l2} & NCE & \ABXbox{26.11} & \ABXbox{15.93} & \ABXbox{12.85} & \ABXbox{10.55} & \ABXbox{9.09} & \ABXbox{3.81}\\
20 & CPC-l4+km50 & \textit{CPC-l2} & NCE & \ABXbox{22.1} & \ABXbox{12.88} & \ABXbox{10.17} & \ABXbox{10.18} & \ABXbox{8.81} & \ABXbox{3.81}\\
9 & CPC-l2+km50 & \textit{CPC-l2} & L1 & \ABXbox{7.3} & \ABXbox{5.25} & \ABXbox{5.71} & \ABXbox{4.89} & \ABXbox{10.96} & \ABXbox{3.81}\\
21 & CPC-l0+km50 & \textit{CPC-l2} & L1 & \ABXbox{26.11} & \ABXbox{13.13} & \ABXbox{10.87} & \ABXbox{9.81} & \ABXbox{15.40} & \ABXbox{3.81}\\
22 & CPC-l4+km50 & \textit{CPC-l2} & L1 & \ABXbox{22.1} & \ABXbox{11.44} & \ABXbox{9.91} & \ABXbox{7.65} & \ABXbox{11.88} & \ABXbox{3.81}\\
10 & CPC-l2+km50 & \textit{CPC-l2} & L2 & \ABXbox{7.3} & \ABXbox{5.42} & \ABXbox{5.09} & \ABXbox{4.20} & \ABXbox{3.68} & \ABXbox{3.81}\\
\midrule
& \multicolumn{9}{l}{\rule{0pt}{2ex} \textit{HuBERT Base Models}} \\
23 & \textit{waveform} & MFCC+km100 & NLL-e & - & \ABXbox{7.13} & \ABXbox{4.18} & \ABXbox{5.03} & \ABXbox{9.05} & \ABXbox{27.88}\\
24 & \textit{waveform} & H1-l6+km500 & NLL-e & - & \ABXbox{6.83} & \ABXbox{4.71} & \ABXbox{4.42} & \ABXbox{3.53} & \ABXbox{6.97}\\
25 & \textit{waveform} & H2-l12+km500 & NLL-e & - & \ABXbox{6.66} & \ABXbox{4.43} & \ABXbox{4.48} & \ABXbox{3.78} & \ABXbox{6.26}\\
\midrule
& \multicolumn{9}{l}{\rule{0pt}{2ex} \textit{BERT Models on HuBERT Discrete Units}} \\
26 & H2-l12+km500 & H2-l12+km500 & NLL-e & \ABXbox{6.26} & \ABXbox{5.32} & \ABXbox{6.51} & \ABXbox{6.74} & \ABXbox{4.43} & \ABXbox{6.26}\\
\midrule

\end{tabular}
\end{adjustbox}
\end{center}
\caption{
Average (within and across) dev-clean ABX error of input features, target features and features from different hidden layers of Transformer model. CPC-lx stands for layer x of CPC-big, Hj-lx stands for layer x of HuBERT j'th iteration.
}
\label{tab:BERTlayersABX}
\end{table*}

\end{document}